\documentclass[reqno,oneside,letterpaper,10pt]{article}
\usepackage[utf8]{inputenc} 
\pdfoutput=1        
\usepackage[textwidth=142mm,textheight=214mm]{geometry}

\usepackage{amsmath,amssymb,amsthm}

\usepackage{microtype}

\usepackage[utf8]{inputenc} 
\usepackage{graphicx} %
\usepackage{subfigure}

\usepackage{algorithm}
\usepackage{algorithmicx}
\usepackage[noend]{algpseudocode} %

\makeatletter
\renewcommand\fs@ruled{\def\@fs@cfont{\bfseries}\let\@fs@capt\floatc@ruled
  \def\@fs@pre{\kern10pt}%
  \def\@fs@post{\kern2pt}%
  \def\@fs@mid{\kern4pt}%
  \let\@fs@iftopcapt\iftrue}
\makeatother

\usepackage[square,sort,comma,numbers]{natbib}
\usepackage{afterpage}
\usepackage[colorlinks,citecolor=blue,urlcolor=blue,linkcolor=blue]{hyperref}

\usepackage{wrapfig}
\usepackage{grffile}
\usepackage{float}
\usepackage{rotating}
\usepackage{enumitem} 
\usepackage{bm}
\usepackage{dsfont}
\usepackage{cite}
\usepackage{color}
\usepackage{soul}
\usepackage{wasysym}
\usepackage{url}

\usepackage[capitalize]{cleveref}  
\usepackage{mathtools}
\usepackage{amsthm}
\usepackage{thmtools}

\usepackage[bf,small]{caption} %

\newcommand{\real}{\mathrm{I\kern-0.175em R}}

\newcommand{\E}{\mathbb E}

\def\bdelta{\bm{\delta}}

\def\btau{\bm{\tau}}

\def\D{\mathcal{D}}

\def\O{\mathcal{O}}

\def\U{\mathcal{U}}

\def\bx{x}

\def\bX{X}

\def\jdash{j^{\prime}}

\def\indicator{\mathds{1}}

\newcommand{\Var}{{\rm Var}} 

\def\Reals{\mathbb{R}}
\renewcommand{\bx}{\bm{x}}
\renewcommand{\bX}{\bm{X}}

\newcommand{\TS}{T}%
\newcommand{\G}{\mathcal{G}}
\newcommand{\defn}[1]{{\bf #1}}

\renewcommand{\min}{\mathsf{min}}
\renewcommand{\max}{\mathsf{max}}

\newcommand{\NS}{\textrm{NSP}}

\newcommand{\PT}{\mathsf{T}}

\newcommand{\samplepartialMPblock}{\mathsf{SampleMondrianBlock}}
\newcommand{\samplepartialMPtree}{\mathsf{SampleMondrianTree}}
\newcommand{\extendMPblock}{\mathsf{ExtendMondrianBlock}}
\newcommand{\extendMPtree}{\mathsf{ExtendMondrianTree}}
\newcommand{\updateCounts}{\mathsf{UpdatePosteriorCounts}}
\newcommand{\initCounts}{\mathsf{InitializePosteriorCounts}}
\newcommand{\predict}{\mathsf{Predict}}
\newcommand{\pnotsplityet}{p_{\mathsf{NotSeparatedYet}}}

\newcommand{\sdim}{\delta}
\newcommand{\bsdim}{\bdelta}
\newcommand{\sloc}{\xi}
\newcommand{\bsloc}{\bm{\xi}}

\newcommand{\bstime}{\btau}

\newcommand{\parent}{\mathsf{parent}}
\newcommand{\child}{\mathsf{child}}
\newcommand{\leaf}[1]{\mathsf{leaves}(#1)}
\newcommand{\nonleaf}[1]{#1 \setminus \leaf {#1}} 
\newcommand{\leafx}[1]{\mathsf{leaf}(#1)}

\newcommand{\ancestors}{\mathsf{ancestors}}
\newcommand{\ancestralpath}{\mathsf{path}}

\newcommand{\dynatree}{\textsf{dynaTree}}
\newcommand{\nonnegative}[1]{\max(#1, 0)}	
\newcommand{\xdomain}{\Reals^D}

\renewcommand{\bell}{\bm{\ell}}
\newcommand{\bu}{\mathbf{\uu}}
\newcommand{\be}{\mathbf{e}}
\newcommand{\bs}{\mathbf{s}}

\renewcommand{\tt}{\tau}
\newcommand{\uu}{u}
\newcommand{\alllabelsequal}[1]{$\mathsf{AllLabelsIdentical}(Y_{N(#1)})$}
\renewcommand{\root}{\epsilon}

\newcommand{\lleft}{\mathsf{left}}
\newcommand{\rright}{\mathsf{right}}
\newcommand{\tab}{\mathsf{tab}}

\newcommand{\leftj}{\mathsf{left}(j)}
\newcommand{\rightj}{\mathsf{right}(j)}
\newcommand{\childj}{\mathsf{child}(j)}

\newcommand{\tj}{\tilde{\jmath}}

\newcommand{\barG}{\overline{G}}

\newcommand{\algcomment}[1]{\Comment{\textit{#1}}}
\algnewcommand{\LineComment}[1]{\(\triangleright\) \textit{#1}}

\newcommand{\figiteronepartition}{2(a)}
\newcommand{\figiteroneapartition}{2(b)}
\newcommand{\figitertwopartition}{2(c)}
\newcommand{\figitertwoapartition}{2(d)}
\newcommand{\figitertwobpartition}{2(e)}
\newcommand{\figiterlastpartition}{2(f)}
\newcommand{\figiteronetree}{2(g)}
\newcommand{\figitertwotree}{2(h)}
\newcommand{\figiterlasttree}{2(i)}

\newcommand{\usps}{\emph{usps}}
\newcommand{\satimage}{\emph{satimages}}
\newcommand{\letter}{\emph{letter}}
\newcommand{\dna}{\emph{dna}}
\newcommand{\Ntrain}{N_{\mathsf{train}}}
\newcommand{\Ntest}{N_{\mathsf{test}}}

\newcommand{\saffari}{{ORF-Saffari}}
\newcommand{\denil}{{ORF-Denil}}

\newcommand{\defas}{:=}

\def\[#1\]{\begin{align}#1\end{align}}

\newcommand{\EE}{\mathbb E}
\newcommand{\MTDIST}{\textrm{MT}}
\newcommand{\MTLAW}[2]{\MTDIST\left(#1, #2\right)}

\newcommand{\kernel}{\textrm{MTx}}
\newcommand{\MTdistribution}[1]{ \MTLAW {\lambda} {\D_{1:{#1}}}}

\newcommand{\xtest}{\bx}
\newcommand{\ytest}{y}
\newcommand{\predictive}[1]{p_{#1}(\ytest|\xtest, \D_{1:N})}

\newcommand{\posterior}[3]{\barG_{#2#3}}

\newlength{\figwidth}
\newlength{\sfigwidth}
\crefname{algocf}{alg.}{algs.}
\Crefname{algocf}{Algorithm}{Algorithms}
\crefname{lem}{Lemma}{Lemmas}
\crefname{prop}{Proposition}{Propositions}
\crefname{cor}{Corollary}{Corollaries}
\crefname{thm}{Theorem}{Theorems}
\crefname{assumption}{Assumption}{Assumptions}

\begin{document}

\title{Mondrian Forests: Efficient Online Random Forests}
\date{}

\author{
Balaji Lakshminarayanan%
\footnote{Corresponding author. Email address: \href{mailto:balaji@gatsby.ucl.ac.uk}{\nolinkurl{balaji@gatsby.ucl.ac.uk}}. }
\\
Gatsby Unit\\University College London
\and
Daniel M. Roy\\
Department of Engineering\\ University of Cambridge
\and
Yee Whye Teh\\
Department of Statistics\\ University of Oxford
}

\newcommand{\fix}{\marginpar{FIX}}
\newcommand{\new}{\marginpar{NEW}}

\maketitle
\thispagestyle{empty} %

\let\thefootnote\relax\footnote{Besides minor corrections and typographical differences, this document is identical in content to, and should be cited as:
B.~Lakshminarayanan, D.~M.~Roy, and Y.~W.~Teh, 
\emph{Mondrian Forests: Efficient Online Random Forests}. 
In,
Z.~Ghahramani and M.~Welling and C.~Cortes and N.~D.~Lawrence and K.~Q.~Weinberger, editors,
\emph{Advances in Neural Information Processing Systems 27} (NIPS), 
pages 3140--3148, 2014.
}

\begin{abstract} 
Ensembles of randomized decision trees, usually referred to as \emph{random forests}, are widely used for classification and regression tasks in machine learning and statistics. Random forests achieve competitive predictive performance and are computationally efficient to train and test, making them excellent candidates for real-world prediction tasks. The most popular random forest variants (such as Breiman's random forest and extremely randomized trees) operate on batches of training data. Online methods are now in greater demand. Existing online random forests, however, require more training data than their batch counterpart to achieve comparable predictive performance. In this work, we use Mondrian processes (Roy and Teh, 2009) to construct ensembles of random decision trees we call \emph{Mondrian forests}. Mondrian forests can be grown in an incremental/online fashion and remarkably, the distribution of online Mondrian forests is the same as that of batch Mondrian forests.  Mondrian forests achieve competitive predictive performance comparable with existing online random forests and periodically re-trained batch random forests, while being more than an order of magnitude faster, thus representing a better computation vs accuracy tradeoff. 
\end{abstract}

\section{Introduction}

Despite being introduced over a decade ago, random forests remain one of the most popular machine learning tools due in part to their accuracy, scalability, and robustness in real-world classification tasks \citep{caruana2006empirical}. (We refer to \citep{criminisi2012decision} for an excellent  %
 survey of random forests.) 
In this paper, we introduce a novel class of random forests---called \emph{Mondrian forests} (MF), due to the fact that the underlying tree structure of each classifier in the ensemble is a so-called \emph{Mondrian process}.  Using the properties of Mondrian processes, we present an efficient \emph{online} algorithm that agrees with its batch counterpart at each iteration.  Not only are online Mondrian forests faster and more accurate than recent proposals for online random forest methods, but they nearly match the accuracy %
 of state-of-the-art \emph{batch} random forest methods trained on the same dataset. %

The paper is organized as follows: In Section~\ref{sec:approach}, we describe our approach at a high-level, and in Sections~\ref{sec:model}, \ref{sec:label distribution}, and \ref{sec:online learning}, we describe the tree structures, label model, and incremental updates/predictions in more detail. 
We discuss related work in Section~\ref{sec:related work},  demonstrate the excellent empirical performance of MF in Section~\ref{sec:experiments}, and conclude in Section~\ref{sec:discussion} with a discussion about future work. 

\section{Approach}\label{sec:approach}

Given $N$ labeled examples $(\bx_1,y_1), \dotsc, (\bx_N,y_N) \in \xdomain \times \mathcal Y$  
as training data, 
our task is to predict labels $y \in \mathcal Y$ for unlabeled test points $\bx \in \xdomain$.  We will focus on multi-class classification where $\mathcal Y\defas \{1,\dotsc,K\}$, %
however, it is possible to extend the methodology to other supervised learning tasks such as regression. 
Let $\bX_{1:n}\defas(\bx_{1},\dotsc,\bx_n)$, $Y_{1:n}\defas(y_1,\dotsc,y_n)$, and $\mathcal D_{1:n}\defas(\bX_{1:n},Y_{1:n})$.

A Mondrian forest classifier is constructed much like a random forest:
Given training data $\mathcal D_{1:N}$,
we sample an independent collection $\TS_1,\dotsc,\TS_M$ of so-called Mondrian trees, which we will describe in the next section. 
The prediction made by each Mondrian tree $\TS_m$ is a distribution $\predictive {\TS_m}$ over the class label $\ytest$ for a test point $\xtest$. 
The prediction made by the Mondrian forest is the average%
$\frac 1 M \sum_{m=1}^M \predictive {\TS_m}$
of the individual tree predictions.
As $M \to \infty$, the average converges at the standard rate to the expectation 
$\EE_{\TS \sim \MTLAW {\lambda} {\D_{1:N}}} [\,\predictive {\TS}]$, %
where $\MTLAW {\lambda} {\D_{1:N}}$ is the distribution of a Mondrian tree.
As the limiting expectation does not depend on $M$, we would not expect to see overfitting behavior as $M$ increases.  A similar observation was made by  Breiman in his seminal article \citep{breiman2001random} introducing random forests.
Note that 
the  averaging procedure above 
 is ensemble model combination 
and \emph{not} Bayesian model averaging.

In the online learning setting, the training examples are presented one after another in a sequence of trials.
Mondrian forests excel in this setting: 
at iteration $N+1$, each Mondrian tree $\TS \sim \MTdistribution{N}$ is updated 
to incorporate the next labeled example $(\bx_{N+1},y_{N+1})$
by sampling an extended tree $\TS'$ from a distribution $\kernel(\lambda, \TS, \D_{N+1})$. 
Using properties of the Mondrian process, we can choose a probability distribution $\kernel$ 
such that $\TS' = \TS$ on $\mathcal D_{1:N}$ 
  and $\TS'$ is distributed according to $\MTdistribution{N+1}$, i.e.,
\[\label{eq:projectivity}
\begin{aligned}
\TS &\sim \MTLAW {\lambda} {\mathcal D_{1:N}}
\\\TS' \mid \TS, \mathcal D_{1:N+1} &\sim \kernel(\lambda, \TS, \mathcal D_{N+1})
\end{aligned}
\qquad \mathit{implies} \qquad \TS' \sim \MTLAW {\lambda} {\mathcal D_{1:N+1}}.
\]
Therefore, the distribution of Mondrian trees trained on a dataset in an incremental fashion is the same as that of Mondrian trees trained on the same dataset in a batch fashion, irrespective of the order in which the data points are observed. To the best of our knowledge, none of the existing online random forests have this property.
Moreover,  we can sample from $\kernel(\lambda, \TS, \mathcal D_{N+1})$ efficiently: 
the complexity scales with the depth of the tree, {which  is typically logarithmic in $N$.} 
 
 While treating the online setting as a sequence of larger and larger batch problems is normally computationally prohibitive, this approach can be achieved efficiently with Mondrian forests. 
In the following sections, we define the Mondrian tree distribution $\MTdistribution{N}$, the label distribution $\predictive {\TS}$, and the update distribution $\kernel(\lambda, \TS, \mathcal D_{N+1})$.

\section{Mondrian trees}\label{sec:model}
\begin{figure}%
\begin{center}
\subfigure[Decision Tree]{
\includegraphics[height=0.9in]{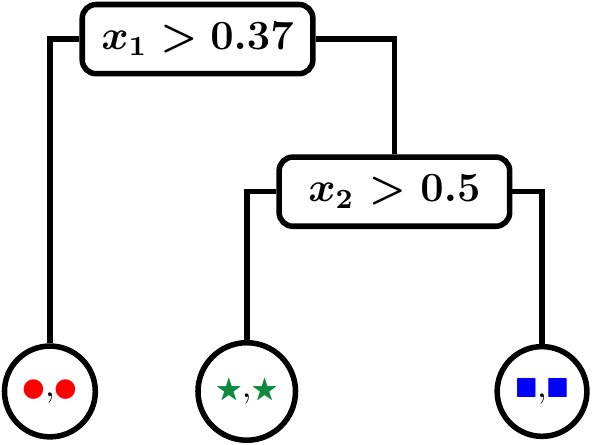}
\includegraphics[height=1.15in]{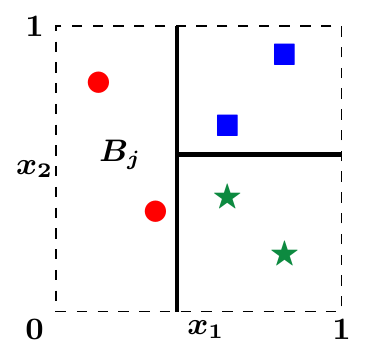}
\label{fig:dtree}
}
\hspace{0.25in}
\subfigure[Mondrian Tree]{
\includegraphics[height=1.16in]{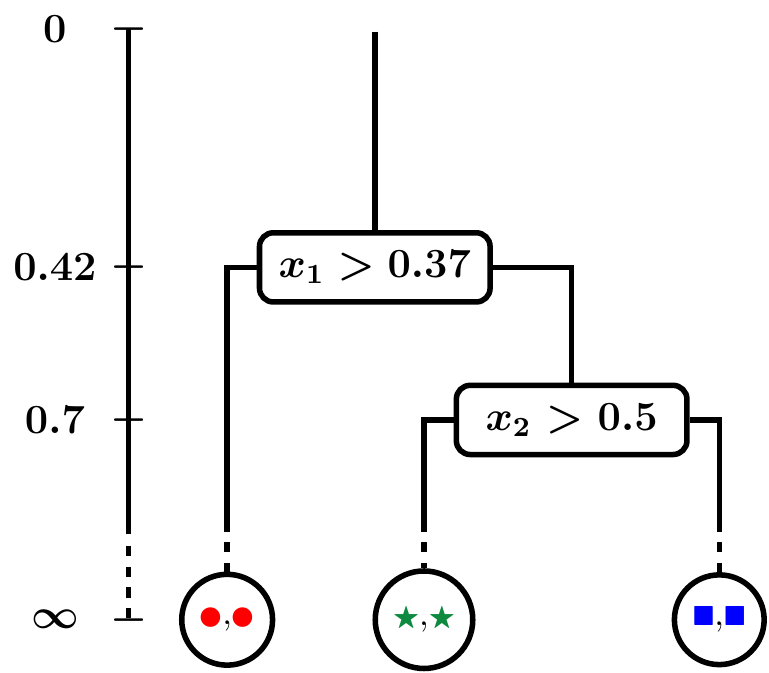}
\includegraphics[height=1.16in]{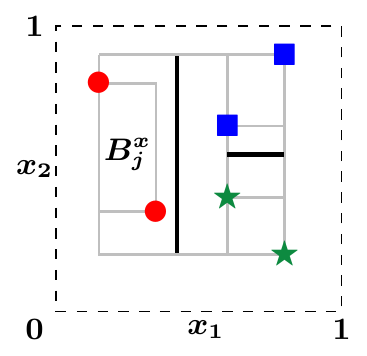}
\label{fig:mtree}
}
\end{center}
\vspace{-0.2in}
\caption{Example of a decision tree in $[0,1]^2$ where $x_1$ and $x_2$ denote horizontal and vertical axis respectively: 
Figure~\ref{fig:dtree} shows tree structure and partition of a decision tree, while Figure~\ref{fig:mtree} shows a Mondrian tree. Note that the Mondrian tree is embedded on a vertical time axis, with each node associated with a time of split and the splits are committed only within the range of the training data in each block (denoted by gray rectangles). %
Let $j$ denote the  left child of the root: $B_j=(0,0.37]\times(0,1]$ denotes the block associated with red circles and $B_j^x\subseteq B_j$ is the smallest rectangle enclosing the two data points.
\\\hspace*{1em}
}
\label{fig:mtreevsdtree}
\end{figure}

For our purposes, a \defn{decision tree} on $\xdomain$ will be a hierarchical, binary partitioning of $\xdomain$ and a rule for predicting the label of test points given training data.
The structure of the decision tree 
is 
a finite, rooted, strictly binary tree $\PT$, i.e., a finite set of \defn{nodes} such that  1) every node $j$ has exactly one \defn{parent} node, except for a distinguished \defn{root} node $\epsilon$ which has no parent, and 2) every node $j$ 
is the parent of exactly zero or two \defn{children} nodes,
called the left child $\leftj$ and the right child $\rightj$.
Denote the leaves of $\PT$ (those nodes without children) by $\leaf\PT$.  Each node of the tree $j\in\PT$  is associated with a block $B_{j}\subset\Reals^D$ of the input space as follows:  At the root, we have $B_{\epsilon}=\Reals^D$, while each \defn{internal} node $j\in\nonleaf\PT$ with two children represents a \defn{split} of its parent's block into two halves, with $\sdim_{j}\in\{1,\dotsc,D\}$ denoting the dimension of the split, and $\sloc_{j}$ denoting the location of the split.  In particular,
\begin{align}
 B_{\leftj} \defas \{\bx\in B_j: x_{\sdim_j}\leq\sloc_j\} \quad \textrm{and} \quad
  B_{\rightj} \defas \{\bx\in B_j: x_{\sdim_j}>\sloc_j\}.
\end{align}
We call the tuple $(\PT,\bsdim,\bsloc)$ a \defn{decision tree}.  
   Note that the blocks associated with the leaves of the tree form a partition of $\Reals^D$.
 We may  write
 $B_j=\bigl(\ell_{j1}, \uu_{j1}\bigr]\times\ldots\times\bigl(\ell_{jD}, \uu_{jD}\bigr]$,
where $\ell_{jd}$ and $u_{jd}$ denote %
the $\ell$ower and $u$pper bounds, %
 respectively, 
of the rectangular block $B_j$ along dimension $d$.
Put $\bell_j=\{\ell_{j1}, \ell_{j2}, \ldots, \ell_{jD}\}$ and $\bu_j=\{\uu_{j1}, \uu_{j2}, \ldots, \uu_{jD}\}$.    
 See
  Figure~\ref{fig:dtree}
   for a simple illustration of a decision tree.

It will be useful to introduce some additional notation. 
Let $\parent(j)$ denote the parent of node $j$. %
Let $N(j)$ denote the indices of training data points at node $j$, i.e., $N(j)=\{n\in\{1,\ldots,N\}: \bx_n\in B_j\}$. Let $\D_{N(j)}=\{\bX_{N(j)}, Y_{N(j)}\}$ denote the features and labels of training data points at node $j$. 
Let $\ell_{jd}^x$ and $u_{jd}^x$ denote the lower and upper bounds %
 of training data points (hence the superscript $x$) respectively in node $j$ along dimension $d$. Let $B_j^x=\bigl(\ell_{j1}^x, \uu_{j1}^x\bigr]\times\ldots\times\bigl(\ell_{jD}^x, \uu_{jD}^x\bigr] \subseteq B_j$ denote the smallest rectangle that encloses the training data points in node $j$.

\newcommand{\Mondrian}{\mathcal M}
\subsection{Mondrian process distribution over decision trees}\label{sec:projectivity}

Mondrian processes, introduced by \citet{RT09}, are families $\{ \Mondrian_t : t \in [0,\infty) \}$ of random, hierarchical binary partitions of $\xdomain$ such that $\Mondrian_t$ is a refinement of $\Mondrian_s$ whenever $t > s$.\footnote{Roy and Teh~\citep{RT09} studied the distribution of $\{ \Mondrian_t : t \le \lambda \}$ and referred to $\lambda$ as the \emph{budget}.
See~\citep[][Chp.~5]{RoyThesis} for more details. We will refer to $t$ as time, not be confused with discrete time in the online learning setting.} 
Mondrian processes are natural candidates for the partition structure of random decision trees, but
Mondrian processes on $\xdomain$ are, in general, infinite structures that we cannot represent all at once.  
Because we only care about the partition on a finite set of observed data, we introduce \defn{Mondrian trees}, which are restrictions of Mondrian processes to a finite set of points.
A Mondrian tree $T$ can be represented by a tuple $(\PT, \bsdim, \bsloc, \bstime)$, where $(\PT,\bsdim,\bsloc)$ is a decision tree and $\bstime=\{\tt_j\}_{j\in\PT}$ associates a time of split $\tt_j \ge 0$ with each node $j$. 
Split times increase with depth, i.e., $\tt_j>\tt_{\parent(j)}$. We abuse notation and define $\tt_{\parent(\root)}=0$.

Given a non-negative \emph{lifetime} parameter $\lambda$ and training data $\D_{1:n}$, 
the generative process for sampling Mondrian trees from $\MTdistribution{n}$ is described in the following two algorithms:
\kern-10pt %
\begin{algorithm}[H]
\caption{$\samplepartialMPtree\bigl(\lambda,\D_{1:n}\bigr)$}
\label{alg:generative process}
\begin{algorithmic}[1]
\State Initialize: $\PT=\emptyset$, $\leaf\PT=\emptyset$, $\bsdim=\emptyset$, $\bsloc=\emptyset$, $\bstime=\emptyset$, $N(\root)=\{1,2,\ldots, n\}$
\State $\samplepartialMPblock\bigl(\root, \D_{N(\root)},\lambda\bigr)$ \algcomment{Algorithm~\ref{alg:generative process:partialblock}}
\end{algorithmic}
\end{algorithm}
\kern-18pt %
\begin{algorithm}[H]
\caption{$\samplepartialMPblock\bigl(j, \D_{N(j)},\lambda\bigr)$}
\label{alg:generative process:partialblock}
\begin{algorithmic}[1]
\State Add $j$ to $\PT$
\State For all $d$, set $\ell_{jd}^x=\min(\bX_{N(j),d}), \uu_{jd}^x=\max(\bX_{N(j),d})$ \algcomment{dimension-wise $\min$ and $\max$}
\State Sample $E$ from exponential distribution with rate $\sum_d (u_{jd}^x-\ell_{jd}^x)$
\If{$\tt_{\parent(j)}+E < \lambda$ } \algcomment{$j$ is an internal node}
\State Set $\tt_j = \tt_{\parent(j)}+E$
\State Sample split dimension $\sdim_j$, choosing $d$ with probability proportional to $u_{jd}^x-\ell_{jd}^x$
\State Sample split location $\sloc_j$ uniformly from interval $[\ell_{j\sdim_j}^x, u_{j\sdim_j}^x]$ %
\State Set $N(\leftj)=\{n\in N(j): \bX_{n,\sdim_j}\leq\sloc_j\}$ and $N(\rightj)=\{n\in N(j): \bX_{n,\sdim_j}>\sloc_j\}$
\State $\samplepartialMPblock\bigl(\leftj,  \D_{N(\leftj)}, \lambda\bigr)$ 
\State $\samplepartialMPblock\bigl(\rightj, \D_{N(\rightj)}, \lambda\bigr)$ 
\Else \algcomment{$j$ is a leaf node}
\State Set $\tt_j=\lambda$ and add $j$ to $\leaf\PT$
\EndIf 
\end{algorithmic}
\end{algorithm}
The procedure starts with the root node $\root$ and recurses down the tree.
In Algorithm~\ref{alg:generative process:partialblock}, we first compute the $\bell_\root^x$ and $\bu_\root^x$ i.e. the lower and upper bounds of $B_\root^x$, the smallest rectangle enclosing $\bX_{N(\root)}$. We sample $E$ from an exponential distribution whose rate is the so-called linear dimension of $B_\root^x$, given by $\sum_d (u_{\root d}^x-\ell_{\root d}^x)$. 
Since $\tt_{\parent(\root)}=0$, $E+\tt_{\parent(\root)}=E$. 
If $E\geq\lambda$, the time of split is not within the lifetime $\lambda$; hence, we assign $\root$ to be a leaf node and the procedure halts. 
(Since $\E[E]=1/\bigl(\sum_d (u_{jd}^x-\ell_{jd}^x)\bigr)$, bigger rectangles are less likely to be leaf nodes.) 
Else, $\root$ is an internal node and we sample a split $(\sdim_\root, \sloc_\root)$ %
 from the \emph{uniform split distribution} on $B_\root^x$. More precisely, we first sample the dimension $\sdim_\root$, taking the value $d$ with probability proportional to $u_{\root d}^x-\ell_{\root d}^x$, and then sample the split location $\sloc_\root$ uniformly from the interval $[\ell_{\root \sdim_\root}^x, u_{\root \sdim_\root}^x]$. The procedure then recurses along the left and right children.

Mondrian trees differ from standard decision trees (e.g. CART, C4.5) in the following ways: (i) the splits are sampled independent of the labels $Y_{N(j)}$;
(ii) every node $j$ is associated with a split time denoted by $\tt_j$;
(iii) the lifetime parameter $\lambda$ controls the total number of splits (similar to the maximum depth parameter for standard decision trees); 
(iv) the split represented by an internal node $j$ holds only within $B_j^x$ and not the whole of $B_j$.  No commitment is made in $B_j \setminus B_j^x$. 
Figure~\ref{fig:mtreevsdtree} illustrates the difference between decision trees and Mondrian trees.

Consider the family of distributions  $\MTLAW {\lambda} {F}$, where $F$ ranges over all possible finite sets of data points.
Due to the fact that these distributions are derived from that of a Mondrian process on $\xdomain$ restricted to 
 a set $F$ of points, the family $\MTLAW {\lambda} {\cdot}$ will be \emph{projective}.  Intuitively, projectivity implies that the tree distributions possess a type of self-consistency. %
  In words, if we sample a Mondrian tree $T$ from $\MTLAW {\lambda} {F}$ and then restrict the tree $T$ to a subset $F' \subseteq F$ of points, then the restricted tree $T'$ has distribution $\MTLAW{\lambda}{F'}$.
Most importantly, projectivity gives us a consistent way to extend a Mondrian tree on a data set $\D_{1:N}$ to a larger data set $\D_{1:N+1}$.
We exploit this property to incrementally grow a Mondrian tree: 
we instantiate the Mondrian tree on the observed training data points; upon observing a new data point $\D_{N+1}$, we \emph{extend} the Mondrian tree by sampling from the conditional distribution of a Mondrian tree on $\D_{1:N+1}$ given its restriction to $\D_{1:N}$, denoted by $\kernel(\lambda, \TS, \mathcal D_{N+1})$ in \eqref{eq:projectivity}.
 Thus, a Mondrian process on $\xdomain$ is represented only where we have observed training data.

\section{Label distribution: model, hierarchical prior, and \\ predictive posterior}\label{sec:label distribution} %

So far, our discussion has been focused on the tree structure. 
In this section, we focus on 
the predictive label distribution, $\predictive {\TS}$, for a tree $\TS = (\PT, \bsdim, \bsloc, \bstime)$, dataset $\D_{1:N}$, and test point~$\bx$.  
Let $\leafx{\bx}$ denote the unique leaf node $j \in \leaf{\PT}$ such that $\bx\in B_{j}$.
Intuitively, we want the predictive label distribution at $\bx$ to be a smoothed version of the empirical distribution of labels for points in $B_{\leafx{\bx}}$ and in $B_{j'}$ for nearby nodes $j'$.
We achieve this smoothing via a hierarchical Bayesian approach:  every node is associated with a label distribution, and a prior is chosen under which the label distribution of a node is similar to that of its parent's.  The predictive $\predictive {\TS}$ is then obtained via marginalization.

As is common in the decision tree literature, we assume the labels within each block are independent of $\bX$ given the tree structure.
For every $j \in \PT$, let $G_j$ denote the distribution of labels at node $j$, %
 and let $\G=\{G_j:j\in \PT\}$ be the set of label distributions at all the nodes in the tree. %
Given $\TS$ and $\G$, the predictive label distribution at $\bx$ is $p(y|\bx, \TS, \G)=G_{\leafx{\bx}}$, i.e., the label distribution at the node $\leafx{\bx}$.
 In this paper, we focus on the case of categorical labels taking values in the set $\{1,\dotsc,K\}$, and so we abuse notation and write 
 $G_{j,k}$ for the probability that a point in $B_j$ is labeled $k$.  %

We model the collection $G_j$, for $j\in \PT$, as a hierarchy of normalized stable processes (NSP) \citep{wood2009stochastic}.
A NSP prior is a distribution over distributions and is a special case of the Pitman-Yor process (PYP) prior where the concentration parameter is taken to zero \citep{pitman2006combinatorial}.\footnote{Taking the discount parameter to zero leads to a Dirichlet process .  Hierarchies of NSPs admit more tractable approximations than hierarchies of Dirichlet processes \citep{wood2009stochastic}, hence our choice here.} The discount parameter $d\in(0,1)$ controls the variation around the base distribution; 
 if $G_j\sim\NS(d, H)$, then $\E[G_{jk}]=H_k$ and $\Var[G_{jk}]=(1-d)H_k(1-H_k)$. %
We use a hierarchical NSP (HNSP) prior over $G_j$ as follows: 
\begin{align}
G_\root|H \sim \NS(d_\root, H),\qquad\textrm{and}\qquad%
G_{j}|G_{\parent(j)} \sim \NS(d_j, G_{\parent(j)}).
\end{align}
This hierarchical prior was first proposed by \citet{wood2009stochastic}.  
Here we take the base distribution $H$ to be the uniform distribution over the $K$ labels,
and set  $d_j=\exp\bigl(-\gamma(\tt_j-\tt_{\parent(j)})\bigr)$.

 Given training data $\D_{1:N}$, the predictive distribution $\predictive {\TS}$ is obtained by integrating over $\G$, i.e.,
 \[
 \predictive {\TS}
 =\E_{\G\sim p_{\TS}(\G|\D_{1:N})}[G_{\leafx{\bx},y}]=\barG_{\leafx{\bx},y},
 \] %
where the posterior over the label distributions is given by 
\[
p_{\TS}(\G|\D_{1:N})\propto p_{\TS}(\G) \prod_{n=1}^N G_{\leafx{\bx_n},y_n}.
\] 
Posterior inference in the HNSP, i.e., computation of the posterior means $\barG_{\leafx{\bx}}$, 
  is a special case of posterior inference in the hierarchical PYP (HPYP). In particular, \citet{teh2006hierarchical} considers the HPYP with multinomial likelihood (in the context of language modeling).  The model considered here is a special case of  \citep{teh2006hierarchical}. Exact inference is intractable and hence we resort to approximations. In particular, we use a fast approximation known as the interpolated Kneser-Ney (IKN) smoothing \citep{teh2006hierarchical}, a popular technique for smoothing probabilities in language modeling \citep{goodman2001bit}. The IKN approximation in \citep{teh2006hierarchical}  can be extended in a straightforward fashion to the online setting, and the computational complexity of adding a new training instance is linear in the depth of the tree. 
 We refer the reader to Appendix~\ref{sec:posterior} for further details.

\section{Online training and prediction} %
\label{sec:online learning}

In this section, we describe the family of distributions $\kernel(\lambda, \TS, \D_{N+1})$, which are used to incrementally add a data point, $\D_{N+1}$, to a tree $\TS$.
These updates are based on the conditional Mondrian algorithm~\citep{RT09}, specialized to a finite set of points. %
In general,
 one or more of the following three operations may be executed %
  while introducing a new data point: (i) introduction of a new split `above' an existing split, (ii) extension of an existing split to the updated extent of the block and (iii) splitting an existing leaf node into two children.
To the best of our knowledge, existing online decision trees use just the third operation, and the first two operations are unique to  Mondrian trees. 
The complete pseudo-code for incrementally updating a Mondrian tree $\TS$ with a new data point $\D$ according to $\kernel(\lambda, \TS,\D)$ is described in the following two algorithms.
Figure~\ref{fig:onlinemondrian} walks through the algorithms on a toy dataset. 
  
\begin{algorithm}[H]%
\caption{$\extendMPtree(\TS,\lambda,\D)$} \label{alg:conditionalmondrian}
\begin{algorithmic}[1]
\State Input: Tree $\TS=(\PT, \bsdim, \bsloc, \bstime)$, new training instance $\D=(\bx, y)$ 
\State $\extendMPblock(\TS,\lambda,\root,\D)$ \algcomment{Algorithm~\ref{alg:extendmondrianblock}}
\end{algorithmic}
\end{algorithm}
\kern-16pt %
\begin{algorithm}[H]%
\caption{$\extendMPblock(\TS,\lambda,j,\D)$} \label{alg:extendmondrianblock}
\begin{algorithmic}[1]
\State Set $\be^\ell = \nonnegative{\bell_j^x - \bx}$ and  $\be^u = \nonnegative{\bx - \bu_j^x}$ \algcomment{$\be^\ell=\be^u=\bm{0}_D$ if $\bx\in B_j^x$}
\State Sample $E$ from exponential distribution with rate $\sum_d (e^\ell_d+e^u_d)$
\If{$\tt_{\parent(j)} + E < \tt_j$}  \algcomment{introduce new parent for node $j$}%
\State Sample split dimension $\sdim$, choosing $d$ with probability proportional to $e^\ell_d+e^u_d$

\State Sample split location $\sloc$ uniformly from interval $[u_{j,\sdim}^x,x_{\sdim}]$ \textbf{if} $x_{\sdim} > u_{j,\sdim}^x$ 
            \textbf{else} $[x_{\sdim},\ell_{j,\sdim}^x]$.

\State Insert a new node $\tj$ just above node $j$ in the tree, and a new leaf $j''$, sibling to $j$, where
\State \qquad $\sdim_{\tj} = \sdim$, $\sloc_{\tj} = \sloc$, 
                       $\tt_{\tj} = \tt_{\parent(j)} + E$, 
                      $\bell_{\tj}^x = \min(\bell_j^x, \bx)$, $\bu_{\tj}^x=\max(\bu_j^x, \bx)$
\State \qquad $j'' = \lleft(\tj)$ \textbf{iff} $x_{\sdim_{\tj}} \le \sloc_{\tj}$ 

\State %
          $\samplepartialMPblock\bigl( j'', \D,  \lambda \bigr)$ %

\Else

\State Update $\bell_j^x \gets \min(\bell_j^x, \bx), \bu_j^x \gets \max(\bu_j^x, \bx)$ \algcomment{update extent of node $j$}%
\If{$j\notin\leaf{\PT}$}  \algcomment{return if $j$ is a leaf node, else recurse down the tree}
\State \textbf{if} $x_{\sdim_j}\leq \sloc_j$ \textbf{then} $\childj=\leftj$ \textbf{else} $\childj=\rightj$
\State $\extendMPblock(\TS,\lambda,\childj,\D)$  \algcomment{recurse on child containing $\D$}
\EndIf

\EndIf
\end{algorithmic}
\end{algorithm}

In practice, random forest implementations stop splitting a node when all the labels are identical and assign it to be a leaf node. To make our MF implementation comparable, we `\emph{pause}' a Mondrian block when all the labels are identical; if a new training instance lies within $B_j$ of a paused leaf node $j$ and has the same label as the rest of the data points in $B_j$, we continue pausing the Mondrian block. We `\emph{un-pause}' the Mondrian block when there is more than one unique label in that block.  Algorithms~\ref{alg:generative process:partialblock:labels} and \ref{alg:extendmondrianblock:labels} in the 
 appendix %
discuss versions of $\samplepartialMPblock$ and $\extendMPblock$  for paused Mondrians. 

\newcommand{\smallR}{R_{ab}}
\newcommand{\bigR}{R_{abc}}
\begin{figure}[htbp]%
\begin{center}
\includegraphics[width=1\columnwidth]{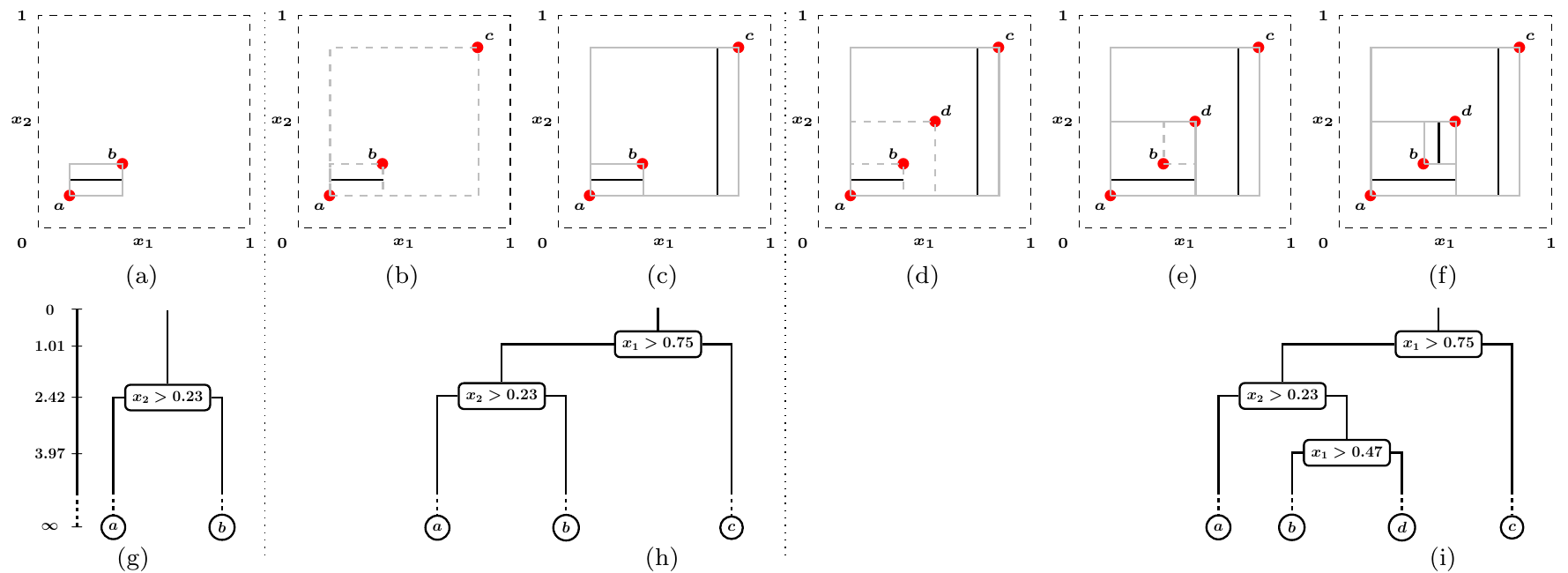}
\vspace{-0.2in}
\caption{Online learning with Mondrian trees on a toy dataset: We assume that $\lambda=\infty, D=2$ and add one data point at each iteration. %
For simplicity, we ignore class labels and denote location of training data with red circles. Figures~\figiteronepartition, \figitertwopartition\ and \figiterlastpartition\ show the partitions after the first, second and third iterations, respectively, with the intermediate figures denoting intermediate steps.
Figures~\figiteronetree, \figitertwotree\ and \figiterlasttree\ show the trees after the first, second and third iterations,  %
along with a shared vertical time axis. 
\\\hspace*{1em}
At iteration 1, we have two training data points, labeled as $a, b$. 
Figures~\figiteronepartition\ and \figiteronetree\  %
show the partition and tree structure of the Mondrian tree.
 Note that even though there is a split $x_2>0.23$ at time $t=2.42$, we {commit this split only within $B_j^x$}  (shown by the gray rectangle). %
\\\hspace*{1em}
At iteration 2, a new data point $c$ is added. Algorithm~\ref{alg:conditionalmondrian} starts with the root node and recurses down the tree. Algorithm~\ref{alg:extendmondrianblock} checks if the new data point lies within $B_\root^x$ by computing the additional extent $\be^\ell$ and $\be^u$.  In this case, $c$ does not lie within $B_\root^x$.   
    Let $\smallR$ and $\bigR$ respectively denote the small gray rectangle (enclosing $a,b$) and big  gray rectangle  (enclosing $a,b,c$) in Figure~\figiteroneapartition.
 While extending the Mondrian from $\smallR$ to $\bigR$, we could either introduce a new split in $\bigR$ outside $\smallR$ or extend the split in $\smallR$ to the new range. To choose between these two options, we sample the time of this new split: we first sample $E$ from an exponential distribution whose rate is the sum of the additional extent, 
 i.e., $\sum_d (e^\ell_d+e^u_d)$, and set the time of the new split to $E+\tt_{\parent(\root)}$. 
If $E+\tt_{\parent(\root)}\leq \tt_\root$, this new split in $\bigR$ can precede the old split in $\smallR$ and a  split is sampled in $\bigR$ outside $\smallR$. In Figures~\figitertwopartition\ and \figitertwotree, $E+\tt_{\parent(\root)}=1.01+0\leq 2.42$, hence a new split  $x_1>0.75$ is introduced. 
The farther a new data point $\bx$ is from $B_j^x$, the higher the rate  $\sum_d (e^\ell_d+e^u_d)$, and subsequently the higher the probability of a new split being introduced, since $\E[E]=1/\bigl(\sum_d (e^\ell_d+e^u_d)\bigr)$.  
A new split in $\bigR$ is sampled  such that %
it is consistent with the existing partition structure in $\smallR$ (i.e., the new split cannot slice through $\smallR$). 
\\\hspace*{1em}
In the final iteration, 
 we add data point $d$. In Figure~\figitertwoapartition, the data point $d$ lies within the extent of the root node, hence we traverse to the left side of the root and update $B_j^x$ of the internal node containing $\{a,b\}$ to include $d$. We could either introduce a new split or extend the split $x_2>0.23$. %
 In Figure~\figitertwobpartition, we extend the split $x_2>0.23$ to the new extent, and traverse to the leaf node in Figure~\figitertwotree\  containing $b$.   In Figures~\figiterlastpartition\ and \figiterlasttree, we sample $E=1.55$ and since $\tt_{\parent(j)}+E=2.42+1.55=3.97\leq\lambda=\infty$, we introduce a new split $x_1>0.47$.
\\\hspace*{1em}
}
\label{fig:onlinemondrian}
\end{center}
\end{figure}

\paragraph{Prediction using Mondrian tree}\label{sec:prediction}
Let $\bx$ denote a test data point. 
If  $\bx$ is already `contained' in the tree $\TS$, i.e., if $\bx \in B_j^x$ for some leaf $j \in \leaf {\PT}$, then the prediction is taken to be $\barG_{\leafx{\bx}}$.  Otherwise, we somehow need to incorporate $\bx$.  One choice is to extend $\TS$ by sampling $\TS'$ from $\kernel(\lambda, \TS,  \bx)$ as described in Algorithm~\ref{alg:conditionalmondrian}, and set the prediction to $\barG_{j}$, where  $j \in \leaf{\PT'}$ is the leaf node containing $\bx$.  A particular extension $\TS'$ might lead to an overly confident prediction; hence, we average over \emph{every} possible extension $\TS'$. 
This integration can be carried out analytically %
 and the computational complexity is linear in the depth of the tree. 
We refer to Appendix~\ref{sec:prediction:app} for further details.

\section{Related work}\label{sec:related work}

The literature on random forests is vast and we do not attempt to cover it comprehensively; we provide a brief review here and refer to \citep{criminisi2012decision} and \citep{denilconsistency} for a recent review of random forests in batch and online settings respectively. 
Classic decision tree induction procedures choose the best split dimension and location from all candidate splits at each node by optimizing some suitable quality criterion (e.g. information gain) in a greedy manner. In a random forest, the individual trees are randomized to de-correlate their predictions. 
The most common strategies for injecting randomness are (i) bagging \citep{breiman1996bagging} %
and (ii) randomly subsampling the set of candidate splits within each node. %

Two popular random forest variants in the batch setting are 
 \emph{Breiman-RF}  \citep{breiman2001random} %
and \emph{Extremely randomized trees (ERT)} \citep{geurts2006extremely}. %
Breiman-RF uses bagging and furthermore, 
at each node, a random $k$-dimensional subset  %
 of the original $D$ features is sampled. %
ERT chooses a $k$ dimensional subset of the features %
 and then chooses one split location each for the $k$ features randomly (unlike Breiman-RF which considers all possible split locations along a dimension). 
  ERT does not use bagging.  
When $k=1$, the ERT trees are \emph{totally randomized} and the splits are chosen independent of the labels; hence the  ERT-$1$ method is very similar to MF in the batch setting in terms of tree induction. (Note that unlike ERT, MF uses HNSP to smooth predictive estimates and allows a test point to branch off into its own node.) Perfect random trees (PERT), proposed by  \citet{cutler2001pert} for classification problems,  produce totally randomized trees similar to ERT-$1$, although there are some slight differences \citep{geurts2006extremely}. %

Existing online random forests %
(\saffari\ \citep{saffari2009line} and \denil\ \citep{denilconsistency}) 
start with an empty tree and grow the tree incrementally. Every leaf of every tree maintains a list of $k$ candidate splits and associated quality scores. When a new data point is added, the scores of the candidate splits at the corresponding leaf node are updated. 
To reduce the risk of choosing a sub-optimal split based on noisy quality scores, additional hyper parameters such as  the minimum number of data points at a leaf node before a decision is made and the minimum threshold for the quality criterion of the best split, are used to assess `confidence' associated with a split.  Once these criteria are satisfied at a leaf node, the best split is chosen (making this node an internal node) and its two children are the new leaf nodes (with their own candidate splits), and the process is repeated.  
These methods could be memory inefficient for deep trees due to the high cost associated with maintaining candidate quality scores for the fringe of potential children  \citep{denilconsistency}. %

There has been some work on incremental induction of decision trees, e.g.~incremental CART \citep{crawford1989extensions}, ITI \citep{utgoff1989incremental}, VFDT %
\citep{domingos2000mining} and dynamic trees \citep{taddy2011},  but to the best of our knowledge, these are focused on learning  decision trees and have not been generalized to online random forests. We do not compare MF to incremental decision trees, since random forests are known to outperform single decision trees. 

Bayesian models of decision trees \citep{chipman1998bayesian,denison1998bayesian} %
 typically specify a distribution over decision trees; such distributions usually depend on $\bX$ and lack the projectivity property of the Mondrian process. More importantly, MF performs ensemble model combination and not Bayesian model averaging over decision trees.  
(See \citep{dietterich2000ensemble} for a discussion on the advantages of ensembles over single models, and \citep{minka2000bayesian} for a comparison of Bayesian model averaging and model combination.)

\section{Empirical evaluation}\label{sec:experiments}

The purpose of these experiments is to evaluate the predictive performance (test accuracy) of MF as a function of (i) fraction of training data and (ii) training time. 
We divide the training data into 100 mini-batches and we compare the performance of online random forests (MF, \saffari\ \citep{saffari2009line}) to batch random forests (Breiman-RF, ERT-$k$, ERT-$1$) which are trained on the same fraction of the training data. %
(We compare MF to dynamic trees as well; see Appendix~\ref{sec:dynamictrees} for more details.)
Our scripts are implemented in Python. %
 We implemented the \saffari\ algorithm as well as ERT in Python for timing comparisons. The scripts can be downloaded from the authors' webpages. We did not implement the \denil\ \citep{denilconsistency} algorithm since the predictive performance reported in  \citep{denilconsistency} is very similar to that of \saffari\ and the computational complexity of the \denil\ algorithm is worse than that of \saffari. We used the Breiman-RF implementation in \emph{scikit-learn} \citep{scikit-learn}.\footnote{The \emph{scikit-learn} implementation uses highly optimized C code, hence we do not compare our runtimes with the  \emph{scikit-learn} implementation. The ERT implementation in \emph{scikit-learn} achieves very similar test accuracy as our ERT implementation, hence we do not report those results here.}

We evaluate on four of the five datasets used in  \citep{saffari2009line} --- we excluded the \emph{mushroom}  dataset as even very simple logical rules achieve $>99\%$ accuracy on this dataset.\footnote{\url{https://archive.ics.uci.edu/ml/machine-learning-databases/mushroom/agaricus-lepiota.names}}  
We re-scaled the datasets such that each feature takes on values in the range $[0,1]$ (by subtracting the $\min$ value along that dimension and dividing by the $\mathsf{range}$ along that dimension, where $\mathsf{range}=\max-\min$).

\begin{figure}[tbp]%
\begin{center}
\includegraphics[width=1.0\columnwidth]{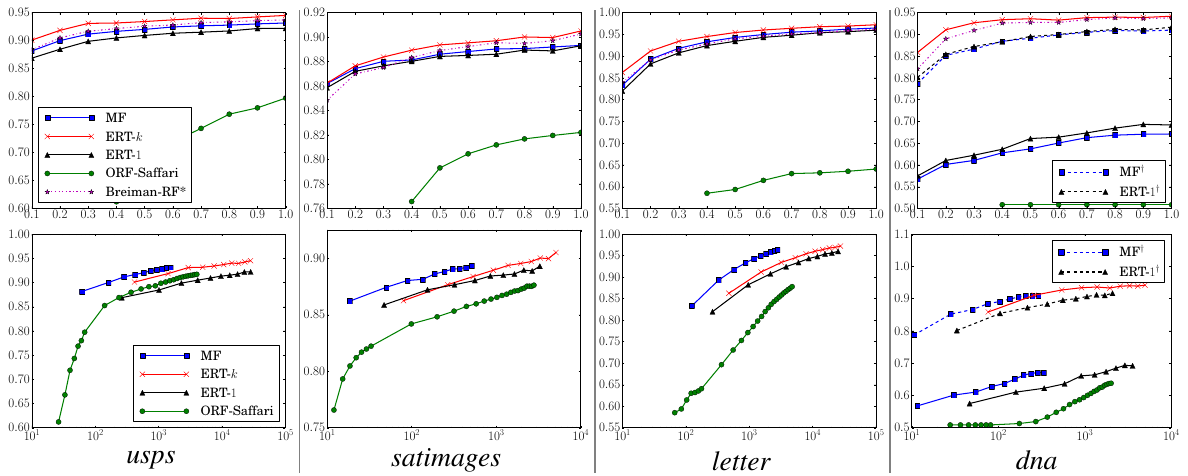}
\end{center}
\caption{Results on various datasets: $y$-axis is test accuracy in both rows. $x$-axis is fraction of training data for the top row and training time (in seconds) for the bottom row. 
We used the pre-defined train/test split.  For \usps\ dataset $D=256, K=10, \Ntrain=7291, \Ntest=2007$; 
for \satimage\ dataset $D=36, K=6, \Ntrain=3104, \Ntest=2000$;  
\letter\ dataset $D=16, K=26, \Ntrain=15000, \Ntest=5000$; 
for \dna\ dataset $D=180, K=3, \Ntrain=1400, \Ntest=1186$. 
}
\label{fig:results}
\end{figure}

As is common in the random forest literature \citep{breiman2001random}, %
we set the number of trees $M=100$. For Mondrian forests, we set the lifetime $\lambda=\infty$ and the HNSP discount parameter $\gamma=10D$. %
 For \saffari, we set $\mathsf{num\_epochs}=20$ (number of passes through the training data) 
 and set the other hyper parameters to the values used in \citep{saffari2009line}. For Breiman-RF and ERT, the hyper parameters are set to default values. We repeat each algorithm with five random initializations and report the mean performance. 
The results are shown in Figure~\ref{fig:results}. (The * in Breiman-RF* indicates \emph{scikit-learn} implementation.)

Comparing test accuracy vs fraction of training data on \usps, \satimage\ and \letter\ datasets, we observe that \textbf{MF achieves accuracy very close to the batch RF versions} 
(Breiman-RF, ERT-$k$, ERT-$1$) trained on the same fraction of the data. %
\textbf{MF significantly outperforms \saffari\ trained on the same fraction of training data}. %
In batch RF versions, the same training data can be used to evaluate candidate splits at a node and its children. However, in the online RF versions (\saffari\ and \denil), incoming training examples are used to evaluate candidate splits just at a current leaf node and new training data are required to evaluate candidate splits every time a new leaf node is created. %
\citet{saffari2009line} recommend multiple passes through the training data to increase the effective number of training samples. 
In a realistic streaming data setup,
 where training examples cannot be stored for multiple passes, MF would require significantly fewer examples than \saffari\ to achieve the same accuracy.

Comparing test accuracy vs training time on \usps, \satimage\ and \letter\ datasets, we observe that \textbf{MF is at least an order of magnitude faster than re-trained batch versions and \saffari}.  For \saffari, we plot test accuracy at the end of every additional   pass; hence it contains additional markers compared to the top row which plots results after a single pass. 
 Re-training batch RF using 100 mini-batches is unfair to MF; in a %
  streaming data setup where the model 
 is updated when a new training instance arrives, 
  MF would be significantly faster than the re-trained batch versions. Assuming trees are balanced after adding each data point, it can be shown that computational cost %
   of MF scales as $\O(N\log N)$ whereas that of re-trained batch RF scales as $\O(N^2\log N)$ (Appendix~\ref{sec:computational complexity}). Appendix~\ref{sec:depth} shows  that the average depth of the forests  trained on above datasets %
    scales as $\O(\log N)$.

It is remarkable that choosing splits independent of labels achieves competitive classification performance. This phenomenon has been observed by others as well---for example, \citet{cutler2001pert} demonstrate that their PERT  classifier (which is similar to batch version of MF) achieves  test accuracy comparable to Breiman-RF on many real world datasets. However, in the presence of irrelevant features, methods which choose splits independent of labels (MF, ERT-$1$) perform worse than Breiman-RF and ERT-$k$ (but still better than \saffari) as indicated by the results on the \dna\ dataset.  We trained MF and ERT-$1$ using just the most relevant 60 attributes amongst the 180 attributes\footnote{\url{https://www.sgi.com/tech/mlc/db/DNA.names}}---these results are indicated as MF$^\dagger$ and ERT-$1^\dagger$ in  Figure~\ref{fig:results}. We observe that, as expected, filtering out irrelevant features significantly improves performance of MF and ERT-$1$.

\vspace{0.1in} %

\section{Discussion}\label{sec:discussion}
We have introduced \emph{Mondrian forests}, a novel class of random forests, which can be trained incrementally in an efficient manner. MF significantly outperforms existing 
 online random forests in terms of training time as well as number of training instances required to achieve a particular test accuracy. Remarkably, MF achieves competitive test accuracy to batch random forests trained on the same fraction of the data. 
MF is unable to handle lots of irrelevant features (since splits are chosen independent of the labels)---one way to use labels to guide splits is via recently proposed Sequential Monte Carlo algorithm for decision trees \citep{LRT13}. The computational complexity of MF is linear in the number of dimensions (since rectangles are represented explicitly) which could be expensive for high dimensional data; we will address this limitation in future work. Random forests have been tremendously influential in machine learning for a variety of tasks; hence lots of other interesting extensions of this work are possible, e.g. MF for regression, theoretical bias-variance analysis of MF, extensions of MF that use hyperplane splits instead of axis-aligned splits.

\subsubsection*{Acknowledgments} %
We would like to thank
Charles Blundell,
Gintare Dziugaite, 
Creighton Heaukulani, 
Jos\'{e} Miguel Hern\'{a}ndez-Lobato,
 Maria Lomeli,
 Alex Smola,  
 Heiko Strathmann and
 Srini Turaga
for helpful discussions and feedback on drafts.
BL gratefully acknowledges generous funding from the Gatsby Charitable Foundation.
This research was carried out in part while DMR held a Research Fellowship at Emmanuel College, Cambridge, with funding also from a Newton International Fellowship through the Royal Society. 
YWT's research leading to these results 
was funded in part by 
the 
European Research Council
 under the European Union's Seventh Framework
Programme (FP7/2007-2013) ERC grant agreement no.~617411.

\bibliography{mondrian-forests,mondrian-forests.handmade}

\begin{thebibliography}{24}
\providecommand{\natexlab}[1]{#1}
\providecommand{\url}[1]{\texttt{#1}}
\expandafter\ifx\csname urlstyle\endcsname\relax
  \providecommand{\doi}[1]{doi: #1}\else
  \providecommand{\doi}{doi: \begingroup \urlstyle{rm}\Url}\fi

\bibitem[Breiman(1996)]{breiman1996bagging}
L.~Breiman.
\newblock Bagging predictors.
\newblock \emph{Mach. Learn.}, 24\penalty0 (2):\penalty0 123--140, 1996.

\bibitem[Breiman(2001)]{breiman2001random}
L.~Breiman.
\newblock Random forests.
\newblock \emph{Mach. Learn.}, 45\penalty0 (1):\penalty0 5--32, 2001.

\bibitem[Caruana and Niculescu-Mizil(2006)]{caruana2006empirical}
R.~Caruana and A.~Niculescu-Mizil.
\newblock An empirical comparison of supervised learning algorithms.
\newblock In \emph{Proc. Int. Conf. Mach. Learn. (ICML)}, 2006.

\bibitem[Chipman et~al.(1998)Chipman, George, and
  McCulloch]{chipman1998bayesian}
H.~A. Chipman, E.~I. George, and R.~E. McCulloch.
\newblock {B}ayesian {CART} model search.
\newblock \emph{J. Am. Stat. Assoc.}, pages 935--948, 1998.

\bibitem[Crawford(1989)]{crawford1989extensions}
S.~L. Crawford.
\newblock Extensions to the {CART} algorithm.
\newblock \emph{Int. J. Man-Machine Stud.}, 31\penalty0 (2):\penalty0 197--217,
  1989.

\bibitem[Criminisi et~al.(2012)Criminisi, Shotton, and
  Konukoglu]{criminisi2012decision}
A.~Criminisi, J.~Shotton, and E.~Konukoglu.
\newblock Decision forests: A unified framework for classification, regression,
  density estimation, manifold learning and semi-supervised learning.
\newblock \emph{Found. Trends Comput. Graphics and Vision}, 7\penalty0
  (2--3):\penalty0 81--227, 2012.

\bibitem[Cutler and Zhao(2001)]{cutler2001pert}
A.~Cutler and G.~Zhao.
\newblock {PERT} - {P}erfect {R}andom {T}ree {E}nsembles.
\newblock \emph{Comput. Sci. and Stat.}, 33:\penalty0 490--497, 2001.

\bibitem[Denil et~al.(2013)Denil, Matheson, and de~Freitas]{denilconsistency}
M.~Denil, D.~Matheson, and N.~de~Freitas.
\newblock Consistency of online random forests.
\newblock In \emph{Proc. Int. Conf. Mach. Learn. (ICML)}, 2013.

\bibitem[Denison et~al.(1998)Denison, Mallick, and Smith]{denison1998bayesian}
D.~G.~T. Denison, B.~K. Mallick, and A.~F.~M. Smith.
\newblock A {B}ayesian {CART} algorithm.
\newblock \emph{Biometrika}, 85\penalty0 (2):\penalty0 363--377, 1998.

\bibitem[Dietterich(2000)]{dietterich2000ensemble}
T.~G. Dietterich.
\newblock Ensemble methods in machine learning.
\newblock In \emph{Multiple classifier systems}, pages 1--15. Springer, 2000.

\bibitem[Domingos and Hulten(2000)]{domingos2000mining}
P.~Domingos and G.~Hulten.
\newblock Mining high-speed data streams.
\newblock In \emph{Proc. 6th ACM SIGKDD Int. Conf. Knowl. Discov. Data Min.
  (KDD)}, pages 71--80. ACM, 2000.

\bibitem[Geurts et~al.(2006)Geurts, Ernst, and Wehenkel]{geurts2006extremely}
P.~Geurts, D.~Ernst, and L.~Wehenkel.
\newblock Extremely randomized trees.
\newblock \emph{Mach. Learn.}, 63\penalty0 (1):\penalty0 3--42, 2006.

\bibitem[Goodman(2001)]{goodman2001bit}
J.~T. Goodman.
\newblock A bit of progress in language modeling.
\newblock \emph{Comput. Speech Lang.}, 15\penalty0 (4):\penalty0 403--434,
  2001.

\bibitem[Lakshminarayanan et~al.(2013)Lakshminarayanan, Roy, and Teh]{LRT13}
B.~Lakshminarayanan, D.~M. Roy, and Y.~W. Teh.
\newblock Top-down particle filtering for {B}ayesian decision trees.
\newblock In \emph{Proc. Int. Conf. Mach. Learn. (ICML)}, 2013.

\bibitem[Minka(2000)]{minka2000bayesian}
T.~P. Minka.
\newblock {B}ayesian model averaging is not model combination.
\newblock MIT Media Lab note.
  \url{http://research.microsoft.com/en-us/um/people/minka/papers/bma.html},
  2000.

\bibitem[Pedregosa et~al.(2011)Pedregosa, Varoquaux, Gramfort, Michel, Thirion,
  Grisel, Blondel, Prettenhofer, Weiss, Dubourg, Vanderplas, Passos,
  Cournapeau, Brucher, Perrot, and Duchesnay]{scikit-learn}
F.~Pedregosa, G.~Varoquaux, A.~Gramfort, V.~Michel, B.~Thirion, O.~Grisel,
  M.~Blondel, P.~Prettenhofer, R.~Weiss, V.~Dubourg, J.~Vanderplas, A.~Passos,
  D.~Cournapeau, M.~Brucher, M.~Perrot, and E.~Duchesnay.
\newblock {Scikit-learn: Machine Learning in Python}.
\newblock \emph{J. Mach. Learn. Res.}, 12:\penalty0 2825--2830, 2011.

\bibitem[Pitman(2006)]{pitman2006combinatorial}
J.~Pitman.
\newblock \emph{Combinatorial stochastic processes}, volume~32.
\newblock Springer, 2006.

\bibitem[Roy(2011)]{RoyThesis}
D.~M. Roy.
\newblock \emph{Computability, inference and modeling in probabilistic
  programming}.
\newblock PhD thesis, Massachusetts Institute of Technology, 2011.
\newblock {\scriptsize\url{http://danroy.org/papers/Roy-PHD-2011.pdf}}.

\bibitem[Roy and Teh(2009)]{RT09}
D.~M. Roy and Y.~W. Teh.
\newblock {The {M}ondrian process}.
\newblock In \emph{Adv. Neural Inform. Proc. Syst. (NIPS)}, volume~21, pages
  27--36, 2009.

\bibitem[Saffari et~al.(2009)Saffari, Leistner, Santner, Godec, and
  Bischof]{saffari2009line}
A.~Saffari, C.~Leistner, J.~Santner, M.~Godec, and H.~Bischof.
\newblock On-line random forests.
\newblock In \emph{Computer Vision Workshops (ICCV Workshops)}. IEEE, 2009.

\bibitem[Taddy et~al.(2011)Taddy, Gramacy, and Polson]{taddy2011}
M.~A. Taddy, R.~B. Gramacy, and N.~G. Polson.
\newblock Dynamic trees for learning and design.
\newblock \emph{J. Am. Stat. Assoc.}, 106\penalty0 (493):\penalty0 109--123,
  2011.

\bibitem[Teh(2006)]{teh2006hierarchical}
Y.~W. Teh.
\newblock A hierarchical {B}ayesian language model based on {P}itman--{Y}or
  processes.
\newblock In \emph{Proc. 21st Int. Conf. on Comp. Ling. and 44th Ann. Meeting
  Assoc. Comp. Ling.}, pages 985--992. Assoc. for Comp. Ling., 2006.

\bibitem[Utgoff(1989)]{utgoff1989incremental}
P.~E. Utgoff.
\newblock Incremental induction of decision trees.
\newblock \emph{Mach. Learn.}, 4\penalty0 (2):\penalty0 161--186, 1989.

\bibitem[Wood et~al.(2009)Wood, Archambeau, Gasthaus, James, and
  Teh]{wood2009stochastic}
F.~Wood, C.~Archambeau, J.~Gasthaus, L.~James, and Y.~W. Teh.
\newblock A stochastic memoizer for sequence data.
\newblock In \emph{Proc. Int. Conf. Mach. Learn. (ICML)}, 2009.

\end{thebibliography}
\bibliographystyle{abbrvnat}

\clearpage
\appendix %
{\centerline{\textbf{\Large{Appendix}}}} %

\section{Posterior inference and prediction using the HNSP}\label{sec:posterior}
Recall that we use a hierarchical Bayesian approach to specify a smooth label distribution $\predictive {\TS}$ for each tree $\TS$. The label prediction at a test point $\bx$ will depend on where $\bx$ falls relative to the existing data in the tree $\TS$.  In this section, we assume that $\bx$ lies within one of the leaf nodes in $\TS$
 , i.e., $\bx\in B_{\leafx{\bx}}^x$, where $\leafx{\bx}\in\leaf{\PT}$. 
 If $\bx$ does not lie within any of the leaf nodes in $\TS$, i.e., $\bx \notin \cup_{j\in\leaf{\PT}} B_j^x$, one could extend the tree  by sampling $\TS'$ from $\kernel(\lambda,\TS, \bx)$, such that $\bx$ lies within a leaf node in $\TS'$ and apply the procedure described below using the extended tree $\TS'$. Appendix~\ref{sec:prediction:app} describes this case in more detail.

Given training data $\D_{1:N}$, a Mondrian tree $\TS$ and the hierarchical prior over $\G$, the predictive label distribution $\predictive {\TS}$ is obtained by integrating over $\G$, i.e.
 \begin{align*}
 \predictive {\TS} 
 &=\E_{\G\sim p_{\TS}(\G|\D_{1:N})}[G_{\leafx{\bx},y}] 
  =\barG_{\leafx{\bx},y}. 
 \end{align*}
Hence, the prediction is given by $\barG_{\leafx{\bx}}$, the posterior mean at $\leafx{\bx}$.  
The posterior mean $\barG_{\leafx{\bx}}$ can be computed using existing techniques, which we review in the rest of this section. %

 Posterior inference in the HNSP is a special case of posterior inference in hierarchical PYP (HPYP). %
 \citet{teh2006hierarchical} considers the HPYP with multinomial likelihood (in the context of language modeling)---the model considered here (HNSP with multinomial likelihood) is a special case of \citep{teh2006hierarchical}. Hence, we just sketch the high level picture and refer the reader to  \citep{teh2006hierarchical} for further details. We first describe posterior inference given $N$ data points $\D_{1:N}$ (batch setting), and later explain how to adapt inference to the online setting. Finally, we describe the computation of the predictive posterior distribution.%

\subsection*{Batch setting}
Posterior inference is done using the Chinese restaurant process %
 representation, wherein every node of the decision tree is a restaurant; the training data points are the customers seated in the tables associated with the leaf node restaurants; these tables are in turn customers at the tables in their corresponding parent level restaurant; the dish served at each table is the class label. Exact inference is intractable and hence we resort to approximations. In particular, we use the approximation known as the interpolated Kneser-Ney (IKN) smoothing, a popular smoothing technique for language modeling \citep{goodman2001bit}. The IKN smoothing can be interpreted as an approximate inference scheme for the HPYP, where the number of tables serving a particular dish in a restaurant is at most one \citep{teh2006hierarchical}. More precisely, if $c_{j,k}$ denotes the number of customers at restaurant $j$ eating dish $k$ and $\tab_{j,k}$ denotes the number of tables at restaurant $j$ serving dish $k$, the IKN approximation sets $\tab_{j,k}=\min(c_{j,k}, 1)$. The counts $c_{j,k}$ and $\tab_{j,k}$ can be computed in a single bottom-up pass as follows: for every leaf node $j\in\leaf\PT$, $c_{j,k}$ is simply the number of  training data points with label $k$ at node $j$; for every internal node $j\in\nonleaf{\PT}$, we set $c_{j,k}=\tab_{\leftj,k}+\tab_{\rightj,k}$. For a leaf node $j$, this procedure is summarized in Algorithm~\ref{alg:ikn:counts:offline}. (Note that this pseudocode just serves as a reference; in practice, these counts are updated in an online fashion, as described in Algorithm~\ref{alg:ikn:counts:online}.)

\begin{algorithm}%
\caption{$\initCounts(j)$}
\label{alg:ikn:counts:offline}
\begin{algorithmic}[1]
\State For all $k$, set $c_{jk} = \#\{n\in N(j): y_n=k\}$
\State Initialize $\jdash=j$
\While{$\mathsf{True}$}
\If{$\jdash\notin\leaf{\PT}$}
\State For all $k$, set $c_{\jdash k}=\tab_{\lleft(\jdash),k} + \tab_{\rright(\jdash),k}$
\EndIf
\State For all $k$, set $\tab_{\jdash k}=\min(c_{\jdash k}, 1)$ \algcomment{IKN approximation}
\If{$\jdash=\root$}
\State \Return
\Else
\State $\jdash\gets\parent(\jdash)$
\EndIf
\EndWhile
\end{algorithmic}
\end{algorithm}

\subsection*{Posterior inference: online setting}
 It is straightforward to extend inference to the online setting. Adding a new data point $\D=(\bx, y)$ affects only the counts along the path from the root to the leaf node of that data point. We update the counts in a bottom-up fashion, starting at the leaf node containing the data point, $\leafx{\bx}$. Due to the nature of the IKN approximation, we can stop at the internal node $j$ where $c_{j,y}=1$ and need not traverse up till the root. This procedure is summarized in Algorithm~\ref{alg:ikn:counts:online}. 
 
 \begin{algorithm}%
\caption{$\updateCounts(j,y)$}
\label{alg:ikn:counts:online}
\begin{algorithmic}[1]
\State $c_{jy} \gets c_{jy} + 1$
\State Initialize $\jdash=j$
\While{$\mathsf{True}$}
\If{$\tab_{\jdash y}=1$} \algcomment{none of the counts above need to be updated}
\State \Return
\Else 
\If{$\jdash\notin\leaf{\PT}$}
\State $c_{\jdash y}=\tab_{\lleft(\jdash),y} + \tab_{\rright(\jdash),y} $
\EndIf
\State $\tab_{\jdash y}=\min(c_{\jdash y}, 1)$ \algcomment{IKN approximation}
\If{$\jdash=\root$}
\State \Return
\Else
\State $\jdash\gets\parent(\jdash)$
\EndIf
\EndIf
\EndWhile
\end{algorithmic}
\end{algorithm}

\paragraph{Predictive posterior computation}
Given the counts $c_{j,k}$ and table assignments $\tab_{j,k}$, the predictive probability (i.e., posterior mean) at node $j$ can be computed recursively as follows: 
\begin{align}
\posterior{\TS}{j}{k}=
\begin{cases} 
\dfrac{c_{j,k}-d_{j}\tab_{j,k}}{c_{j,\cdot}} + \dfrac{d_{j} \tab_{j,\cdot}}{c_{j,\cdot}} \ \posterior{\TS}{\parent(j),}{k} 
&{c_{j,\cdot} > 0},\\
\posterior{\TS}{\parent(j),}{k} & {c_{j,\cdot} = 0},
\end{cases}
\label{eq:prediction}
\end{align}
where $c_{j,\cdot}=\sum_k c_{j,k}$, $\tab_{j,\cdot}=\sum_k \tab_{j,k}$, and 
 $d_j \defas \exp \bigl( -\gamma(\tt_{j}-\tt_{\parent(j)}) \bigr)$ is the \emph{discount} for node $j$,  defined in Section \ref{sec:label distribution}.  Informally, the discount interpolates between the counts $c$ and the prior. If the discount 
 $d_j \approx 1$, then $\barG_j$ is more like its parent $\barG_{\parent(j)}$. If $d_j \approx 0$, then $\barG_j$ weights the counts more.  
These predictive probabilities can be computed in a single top-down pass as shown in Algorithm~\ref{alg:compute:posterior:predictive}. %

\begin{algorithm}%
\caption{$\mathsf{ComputePosteriorPredictiveDistribution}\bigl(\TS,\G\bigr)$}
\label{alg:compute:posterior:predictive}
\begin{algorithmic}[1]
\State \LineComment{Description of top-down pass to compute posterior predictive distribution given by (\ref{eq:prediction})}
\State \LineComment{$\barG_{jk}$ denotes the %
posterior probability of $y=k$ at node $j$ }
\State Initialize the ordered set $J=\{\root\}$ 
\While{$J$ not empty}
\State Pop the first element of $J$ 
\If{$j=\root$}
\State $\posterior{\TS}{\parent(\root)}{}=H$
\EndIf
\State Set $d=\exp\bigl(-\gamma (\tt_{j}-\tt_{\parent(j)})\bigr)$
\State For all $k$, set $\barG_{jk}=c_{j,\cdot}^{-1}\Bigl(c_{j,k}-{d}\ \tab_{j,k}+ {d}\ \tab_{j,\cdot} \ %
\posterior{\TS}{\parent(j),}{k}
\Bigr)$
\If{$j\notin\leaf{\PT}$}
 \State Append $\leftj$ and $\rightj$ to the end of the ordered set $J$
\EndIf
\EndWhile
\end{algorithmic}
\end{algorithm}

\section{Prediction using Mondrian tree}\label{sec:prediction:app}

Let $\bx$ denote a test data point. We are interested in the predictive probability of $y$ at $\bx$, denoted by $\predictive{\TS}$. %
As in typical decision trees, the process involves a top-down tree traversal, starting from the root. 
If  $\bx$ is already `contained' in the tree $\TS$, i.e., if $\bx \in B_j^x$ for some leaf $j \in \leaf {\PT}$, then the prediction is taken to be $\barG_{\leafx{\bx}}$, which is computed as described in Appendix~\ref{sec:posterior}. Otherwise, we somehow need to incorporate $\bx$.  One choice is to extend $\TS$ by sampling $\TS'$ from $\kernel(\lambda, \TS, \bx)$ as described in Algorithm~\ref{alg:conditionalmondrian}, and set the prediction to $\barG_{j}$, where  $j \in \leaf{\PT'}$ is the leaf node containing $\bx$.  A particular extension $\TS'$ might lead to an overly confident prediction; hence, we average over \emph{every} possible extension $\TS'$. 
This expectation can be carried out analytically, using properties of the Mondrian process, as we show below.

Let $\ancestors(j)$ denote the set of all ancestors of node $j$. Let $\ancestralpath(j) = \{j\} \cup \ancestors(j)$, that is, the set of all nodes along the ancestral path from $j$ to the root.  Recall that $\leafx{\bx}$ is the unique leaf node in $\PT$ such that $\bx\in B_{\leafx{\bx}}$. If the test point $\bx\in B_{\leafx{\bx}}^x$ (i.e., $\bx$ lies within the `gray rectangle' at the leaf node), it can never branch off; else, it can branch off at one or more points along the path from the root to $\leafx{\bx}$. More precisely, if $\bx$ lies outside $B_j^x$ at node $j$, the probability that $\bx$ will branch off into its own node at node $j$, denoted by\footnote{The superscript $s$ in $p^{s}_j(\bx)$ is used to denote the fact that this split `separates' the test data point $\bx$ into its own leaf node.} $p^{s}_j(\bx)$, is equal to the probability that a split exists in  $B_j$ outside $B_j^x$, which is %
\begin{align*}
p^{s}_j(\bx)=1-\exp\bigl(-\Delta_j\eta_j(\bx)\bigr), \quad \textrm{where } 
\eta_j(\bx)=\sum_d\bigl(\nonnegative{x_d-u_{jd}^x}+\nonnegative{\ell_{jd}^x-x_d}\bigr),
\end{align*}
and $\Delta_j=\tt_j-\tt_{\parent(j)}$. Note that $p^{s}_j(\bx)=0$ if $\bx$ lies within $B_j^x$ (i.e., if $\ell_{jd}^x\le x_d\le u_{jd}^x$ for all $d$).  The probability of $\bx$ not branching off before reaching node $j$ is given by $\prod_{\jdash\in\ancestors(j)} (1-p^{s}_{\jdash}(\bx))$. 

If $\bx\in B_{\leafx{\bx}}^x$, the prediction is given by $\barG_{\leafx{\bx}}$.  If there is a split in $B_j$ outside $B_j^x$, let $\tj$ denote the new parent of $j$ and $\child(\tj)$ denote the child node containing just the test data point,; in this case, the prediction is $\barG_{\child(\tj)}$. Averaging over the location where the test point branches off, we obtain 
\begin{align}
\predictive{\TS}=& %
\sum_{j\in\ancestralpath(\leafx{\bx})} \Bigl(\prod_{\jdash\in\ancestors(j)} (1-p^{s}_{\jdash}(\bx))\Bigr)F_j(\bx),
\label{eq:pygivenx_tree}
\end{align}
where
\begin{align}
F_j(\bx)=p^s_j(\bx)\E_{\Delta_{\tj}}\Bigl[\posterior{\TS}{\child(\tj)}{}\Bigr]+ \indicator[j=\leafx{\bx}] (1-p^s_j(\bx)) \posterior{\TS}{\leafx{\bx}}{}.
\label{eq:f}
\end{align}
 The second term in $F_j(\bx)$ needs to be computed only for the leaf node $\leafx{\bx}$ and is simply the posterior mean of $G_{\leafx{\bx}}$ weighted by $1-p^s_{\leafx{\bx}}(\bx)$. 
 The posterior mean of $G_{\leafx{x}}$, given by $\barG_{\leafx{\bx}}$, %
 can be computed using (\ref{eq:prediction}).  The first term in $F_j(\bx)$ is simply the posterior mean of $G_{\child(\tj)}$, averaged over $\Delta_{\tj}$, weighted by $p^s_j(\bx)$.  
 Since no labels are observed in  $\child(\tj)$, $c_{\child(\tj),\cdot}=0$, hence from (\ref{eq:prediction}), we have $\barG_{\child(\tj)}=\barG_{\tj}$. We compute  $\barG_{\tj}$ using (\ref{eq:prediction}). 
 We average over $\Delta_{\tj}$ due to the fact that the discount in (\ref{eq:prediction}) for the node $\tj$ depends on $\tt_{\tj}-\tt_{\parent(\tj)}=\Delta_{\tj}$. To average over all valid split times $\tt_{\tj}$, we compute expectation w.r.t.~$\Delta_{\tj}$ which is distributed according to a truncated exponential with rate $\eta_j(\bx)$, truncated to the interval $[0,\Delta_j]$.  

The procedure for computing $\predictive{\TS}$ for any $\bx\in\Reals^D$ is summarized in Algorithm~\ref{alg:predict}. %
The predictive probability assigned by a Mondrian forest is the average of the predictive probability of the $M$ trees, i.e., $\frac{1}{M}\sum_m \predictive{\TS_m}$. 

\begin{algorithm}%
\caption{$\predict\bigl(\TS,\bx\bigr)$}
\label{alg:predict}
\begin{algorithmic}[1]
\State \LineComment{Description of prediction using a Mondrian tree, given by (\ref{eq:pygivenx_tree})} 
\State Initialize $j=\root$ and $\pnotsplityet=1$
\State Initialize $\bs = \bm{0}_K$ \algcomment{$\bs$ is $K$-dimensional vector where $s_k=%
 p_{\TS}(y=k|\bx, \D_{1:N})$}
\While{$\mathsf{True}$}
\State Set  $\Delta_j=\tt_j-\tt_{\parent(j)}$ and $\eta_j(\bx)=\sum_d\bigl(\nonnegative{x_d-u_{jd}^x}+\nonnegative{\ell_{jd}^x-x_d}\bigr)$
\State Set $p^{s}_j(\bx)=1-\exp\bigl(-\Delta_j\eta_j(\bx)\bigr)$
\If{$p^{s}_j(\bx)>0$}
\State \LineComment{Let $\bx$ branch off into its own node $\child(\tj)$, creating a new node $\tj$ which is the parent of $j$ and $\child(\tj)$. $\barG_{\child(\tj)}=\barG_{\tj}$ from (\ref{eq:prediction}) since $c_{\child(\tj),\cdot}=0$.}
\State Compute expected discount $\bar{d}=\E_{\Delta}[\exp(-\gamma\Delta)]$ where $\Delta$ is drawn from a truncated exponential with rate $\eta_j(\bx)$, truncated to the interval $[0,\Delta_j]$.
\State For all $k$, set $c_{\tj,k}=\tab_{\tj,k}=\min(c_{j,k}, 1)$
\State For all $k$, set $\barG_{\tj k}=c_{\tj,\cdot}^{-1}\Bigl(c_{\tj,k}-\bar{d}\ \tab_{\tj,k}+ \bar{d}\ \tab_{\tj,\cdot} \ \barG_{\parent(\tj),k}\Bigr)$ \algcomment{Algorithm~\ref{alg:compute:posterior:predictive} and (\ref{eq:f})}
\State For all $k$, update $s_k\gets s_k +  \pnotsplityet  \ p^{s}_j(\bx) \barG_{\tj k}$
\EndIf

\If{$j\in\leaf{\PT}$}
\State For all $k$, update $s_k\gets s_k +  \pnotsplityet (1-p^s_j(\bx)) \barG_{jk}$\algcomment{Algorithm~\ref{alg:compute:posterior:predictive} and (\ref{eq:f})}
\State \Return predictive probability $\bs$ where $s_k=p_{\TS}(y=k|\bx, \D_{1:N})$
\Else
\State $\pnotsplityet \gets \pnotsplityet (1-p^{s}_j(\bx))$
\State \textbf{if} $x_{\sdim_j}\leq\sloc_j$ \textbf{then} $j\gets\leftj$ \textbf{else} $j\gets\rightj$ \algcomment{recurse to the child where $\bx$ lies}
\EndIf
\EndWhile
\end{algorithmic}
\end{algorithm}

\section{Computational complexity}\label{sec:computational complexity}
We discuss the computational complexity associated with a single Mondrian tree. The complexity of a forest is simply $M$ times that of a single tree; however, this computation can be trivially parallelized since there is no interaction between the trees. Assume that the $N$ data points are processed one by one. Assuming the data points form a balanced binary tree after each update, the computational cost of processing the $n^{th}$ data point is at most $\O(\log n)$ (add the data point into its own leaf, update posterior counts for HNSP in bottom-up pass from leaf to root). %
The overall cost to process $N$ data points is $\O(\sum_{n=1}^N \log n)=\O(\log N!)$, which for large $N$ tends to $\O(N \log N)$ (using Stirling approximation for the factorial function). %
For offline RF and ERT, the expected complexity with $n$ data points is $\O(n\log n)$. The complexity of the re-trained version is  $\O(\sum_{n=1}^N n\log n)=\O(\log \prod_{n=1}^N n^n)$, which for large $N$ tends to $\O(N^2\log N)$ (using asymptotic expansion of the hyper factorial function).%

\section{Pseudocode for paused Mondrians}
In this section, we discuss versions of $\samplepartialMPblock$ and $\extendMPblock$  for paused Mondrians. For completeness, we also provide the updates necessary for the  IKN approximation.

\begin{algorithm}%
\caption{$\samplepartialMPblock\bigl(j, \D_{N(j)},\lambda\bigr)$ version that depends on labels}
\label{alg:generative process:partialblock:labels}
\begin{algorithmic}[1]
\State Add $j$ to $\PT$
\State For all $d$, set $\ell_{jd}^x=\min(\bX_{N(j),d}), \uu_{jd}^x=\max(\bX_{N(j),d})$ \algcomment{dimension-wise $\min$ and $\max$}
\If{\alllabelsequal{j}}
\State Set $\tt_j=\lambda$ \algcomment{pause Mondrian}
\Else
\State Sample $E$ from exponential distribution with rate $\sum_d (u_{jd}^x-\ell_{jd}^x)$
\State Set $\tt_j=\tt_{\parent(j)}+E$
\EndIf
\If{$\tt_j < \lambda$ }
\State Sample split dimension $\sdim_j$ with probability of choosing $d$ proportional to $u_{jd}^x-\ell_{jd}^x$
\State Sample split location $\sloc_j$ along dimension $\sdim_j$ from an uniform distribution over $\U[\ell_{jd}^x, u_{jd}^x]$
\State Set $N(\leftj)=\{n\in N(j): \bX_{n,\sdim_j}\leq\sloc_j\}$ and $N(\rightj)=\{n\in N(j): \bX_{n,\sdim_j}>\sloc_j\}$
\State $\samplepartialMPblock\bigl(\leftj,  \D_{N(\leftj)}, \lambda\bigr)$ 
\State $\samplepartialMPblock\bigl(\rightj, \D_{N(\rightj)}, \lambda\bigr)$ 
\Else
\State Set $\tt_j=\lambda$ and add $j$ to $\leaf\PT$ \algcomment{$j$ is a leaf node}
\State $\initCounts(j)$ \algcomment{Algorithm~\ref{alg:ikn:counts:offline}}
\EndIf 
\end{algorithmic}
\end{algorithm}

\begin{algorithm}[t]%
\caption{$\extendMPblock(\TS,\lambda,j,\D)$ version that depends on labels} \label{alg:extendmondrianblock:labels}
\begin{algorithmic}[1]
\If{\alllabelsequal{j}} \algcomment{paused Mondrian leaf}
\State Update extent $\bell_j^x\gets\min(\bell_j^x, \bx), \bu_j^x\gets\max(\bu_j^x, \bx)$
\State Append $\D$ to $\D_{N(j)}$ \algcomment{append $\bx$ to $X_{N(j)}$ and $y$ to $Y_{N(j)}$}  %
\If{$y=\mathsf{unique}(Y_{N(j)})$}
\State $\updateCounts(j,y)$ \algcomment{Algorithm~\ref{alg:ikn:counts:online}}
\State \Return \algcomment{continue pausing}
\Else
 \State Remove $j$ from $\leaf{\PT}$
 \State $\samplepartialMPblock\bigl(j,  \D_{N(j)}, \lambda\bigr)$ \algcomment{un-pause Mondrian}
\EndIf
\Else
\State Set $\be^\ell = \nonnegative{\bell_j^x - \bx}$ and  $\be^u = \nonnegative{\bx - \bu_j^x}$ \algcomment{$\be^\ell=\be^u=\bm{0}_D$ if $\bx\in B_j^x$}
\State Sample $E$ from exponential distribution with rate $\sum_d (e^\ell_d+e^u_d)$
\If{$\tt_{\parent(j)}+E < \tt_j$}  \algcomment{introduce new parent for node $j$}%
\State Create new Mondrian block $\tj$ where $\bell_{\tj}^x = \min(\bell_j^x, \bx)$ and $\bu_{\tj}^x=\max(\bu_j^x, \bx)$
\State Sample $\sdim_{\tj}$ with $\Pr(\sdim_{\tj}=d)$ proportional to $e^\ell_d+e^u_d$
\State \textbf{if} $x_{\sdim_{\tj}} > u_{j,\sdim_{\tj}}^x$, \textbf{then} sample $\sloc_{\tj}$ from $\U[u_{j,\sdim_{\tj}}^x,x_{\sdim_{\tj}}]$, \textbf{else} sample $\sloc_{\tj}$ from $\U([x_{\sdim_{\tj}},\ell_{j,\sdim_{\tj}}^x])$%
\If{$j=\root$} \algcomment{set $\tj$ as the new root}
\State $\root\gets\tj$
\Else \algcomment{set $\tj$ as child of $\parent(j)$}
\State \textbf{if} $j=\lleft(\parent(j))$,  \textbf{then}  $\lleft(\parent(j))\gets\tj$, \textbf{else}  $\rright(\parent(j))\gets\tj$ %
\EndIf
\If{$x_{\sdim_{\tj}} > \sloc_{\tj}$}%
\State Set $\lleft(\tj) = j$
and %
$\samplepartialMPblock\bigl(\rright(\tj), \D,  \lambda\bigr)$ \algcomment{create new leaf for $x$}
\Else
\State Set $\rright(\tj)=j$ 
and %
 $\samplepartialMPblock\bigl(\lleft(\tj),  \D,  \lambda\bigr)$  \algcomment{create new leaf for $x$}
\EndIf

\Else

\State Update $\bell_j^x \gets \min(\bell_j^x, \bx), \bu_j^x \gets \max(\bu_j^x, \bx)$ \algcomment{update extent of node $j$}%
\If{$j\notin\leaf{\PT}$}  \algcomment{return if $j$ is a leaf node, else recurse down the tree}
\State \textbf{if} $x_{\sdim_j}\leq \sloc_j$ \textbf{then} $\childj=\leftj$ \textbf{else} $\childj=\rightj$
\State $\extendMPblock(\TS,\lambda,\childj,\D)$  \algcomment{recurse on child containing $x$}
\EndIf
\EndIf

\EndIf
\end{algorithmic}
\end{algorithm}

\newpage
\section{Depth of trees}\label{sec:depth}
We computed the average depth of the trees in the forest, where depth of a leaf node is weighted by fraction of data points at that leaf node. The hyper-parameter settings and experimental setup are described in Section~\ref{sec:experiments}. Table~\ref{tab:depth} reports the average depth (and standard deviations) for Mondrian forests trained on different datasets. The values suggest that the depth of the forest scales as $\log N$ rather than $N$. %

\begin{table}[htdp]
\begin{center}
\begin{tabular}{|c|c|c|c|}
\hline
Dataset & $\Ntrain$ & $\log_2 \Ntrain$ & depth \\ \hline
\usps &  7291 & 12.8 & 19.1 $\pm$ 1.3 \\
\satimage & 3104 & 11.6 & 17.4 $\pm$ 1.6\\
\letter & 15000 & 13.9 & 23.2 $\pm$ 1.8 \\
\dna & 1400 & 10.5 &12.0 $\pm$ 0.3 \\
\hline
\end{tabular}
\end{center}
\caption{Average depth of  Mondrian forests trained on different datasets.}
\label{tab:depth}
\end{table}%

\section{Comparison to dynamic trees}\label{sec:dynamictrees}
Dynamic trees \citep{taddy2011} approximate the Bayesian posterior over decision trees in an online fashion. Specifically, dynamic trees maintain a particle approximation to the true posterior; the prediction at a test point is a weighted average of the predictions made by the individual particles. While this averaging procedure appears similar to online random forests at first sight, there is a key difference: MF (and other random forests) performs ensemble model combination whereas dynamic trees use Bayesian model averaging. In the limit of infinite data, the Bayesian posterior would converge to a single tree \citep{minka2000bayesian}, whereas MF would still average predictions over multiple trees. Hence, we expect MF to outperform dynamic trees in scenarios where a single decision tree is insufficient to explain the data.

\begin{figure}[htbp]%
\begin{center}
\includegraphics[width=1.0\columnwidth]{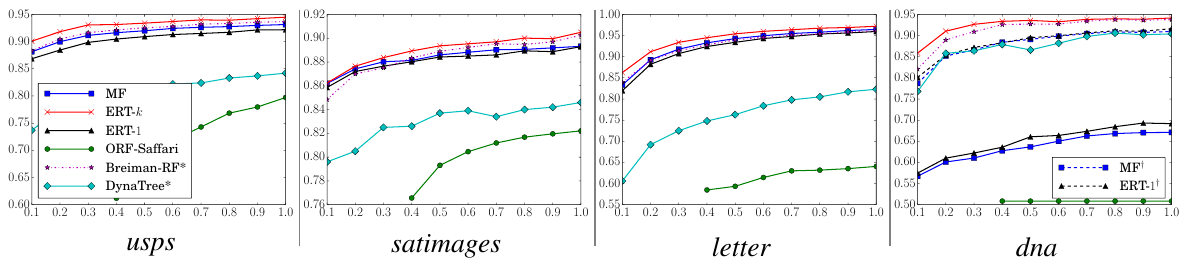}
\end{center}
\caption{Results on various datasets: $y$-axis is test accuracy in both rows. $x$-axis is fraction of training data. The setup is identical to that of Fig.~\ref{fig:results}. MF achieves significantly higher test accuracies than dynamic trees on \usps, \satimage\ and \letter\ datasets and MF$^\dagger$ achieves similar test accuracy as dynamic trees on the \dna\ dataset.}
\label{fig:results:dynatree}
\end{figure}

To experimentally validate our hypothesis, we %
evaluate the empirical performance of dynamic trees  using the \dynatree\footnote{\url{http://cran.r-project.org/web/packages/dynaTree/index.html}} R package provided by the authors of the paper.  Note that while dynamic trees can use `linear leaves' (strong since prediction at a leaf depends on X) or `constant leaves' for regression tasks, they use `multinomial leaves' for classification tasks which corresponds to a `weak learner'. We set the number of particles to 100 (equals the number of trees used in MF) and the number of passes, $R=2$ (their code does not support $R=1$) and set the remaining parameters to their default values. Fig.~\ref{fig:results:dynatree} compares the performance of dynamic trees to MF and other random forest variants. (The performance of all methods other than dynamic trees is identical to that of Fig.~\ref{fig:results}.)

\vspace{0.195in}
\vspace{0.195in}

We observe that  MF achieves significantly higher test accuracies than dynamic trees on \usps, \satimage\ and \letter\ datasets.
On \dna\ dataset, dynamic trees outperform MF (indicating the usefulness of using labels to guide splits) --- however, MF with feature selection (MF$^\dagger$) achieves similar performance as dynamic trees. 
 All the batch random forest methods are superior to dynamic trees which suggests that decision trees are not sufficient to explain these real world datasets and that model combination is helpful.

\end{document}